Reliability of PET/CT shape and heterogeneity features in functional and morphological components of Non-Small Cell Lung Cancer tumors: a repeatability analysis in a prospective multi-center cohort

Marie-Charlotte Desseroit[1,2], MSc., Florent Tixier[2], PhD, Wolfgang A. Weber[3], MD, PhD, Barry A. Siegel[4], MD, Catherine Cheze Le Rest[2], MD, PhD, Dimitris Visvikis[1], PhD, Mathieu Hatt[1], PhD

[1] INSERM, UMR 1101, LaTIM, University of Brest, IBSAM, Brest, France.

[2] Academic department of nuclear Medicine, CHU Milétrie, Poitiers, France.

[3] Memorial Sloan Kettering Cancer Center, New-York, New-York.

[4] Mallinckrodt Institute of Radiology and the Siteman Cancer Center, Washington University School of Medicine, St. Louis, Missouri.

**Corresponding author:**

Marie-Charlotte Desseroit INSERM, UMR 1101, LaTIM
CHRU Morvan, 2 avenue Foch
29609, Brest, France
Tel: +33(0)2.98.01.81.75 - Fax: +33(0)2.98.01.81.24
e-mail: Marie-Charlotte.Desseroit@etudiant.univ-brest.fr

**Wordcount:** ~4880

**Disclosure of Conflicts of Interest:** No potential conflicts of interest.

**Funding**: M-C Desseroit's PhD is partly funded by Brest Métropôle Océane. F. Tixier is funded by the association "Sport and Collection", CHRU Poitiers. This work has received a French government support granted to the CominLabs excellence laboratory and managed by the National Research Agency in the "Investing for the Future" program under reference ANR-10-LABX-07-01. With the support of the National Institute of Cancer (INCa project #C14020NS). The original trials from which the images used in this study were obtained were supported by the U.S. National Cancer Institute through grants U01-CA079778 and U01-CA080098 and by Merck & Co., Inc.

**Short title**: PET/CT texture features repeatability

**ABSTRACT**

Purpose: The main purpose of this study was to assess the reliability of shape and heterogeneity features in both Positron Emission Tomography (PET) and low-dose Computed Tomography (CT) components of PET/CT. A secondary objective was to investigate the impact of image quantization.

Material and methods: A Health Insurance Portability and Accountability Act -compliant secondary analysis of deidentified prospectively acquired PET/CT test-retest datasets of 74 patients from multi-center Merck and ACRIN trials was performed. Metabolically active volumes were automatically delineated on PET with Fuzzy Locally Adaptive Bayesian algorithm. 3DSlicer$^{TM}$ was used to semi-automatically delineate the anatomical volumes on low-dose CT components. Two quantization methods were considered: a quantization into a set number of bins (quantization$_B$) and an alternative quantization with bins of fixed width (quantization$_W$). Four shape descriptors, ten first-order metrics and 26 textural features were computed. Bland-Altman analysis was used to quantify repeatability. Features were subsequently categorized as very reliable, reliable, moderately reliable and poorly reliable with respect to the corresponding volume variability.

Results: Repeatability was highly variable amongst features. Numerous metrics were identified as poorly or moderately reliable. Others were (very) reliable in both modalities, and in all categories (shape, $1^{st}$-, $2^{nd}$- and $3^{rd}$-order metrics). Image quantization played a major role in the features repeatability. Features were more reliable in PET with quantization$_B$, whereas quantization$_W$ showed better results in CT.

Conclusion: The test-retest repeatability of shape and heterogeneity features in PET and low-dose CT varied greatly amongst metrics. The level of repeatability also depended strongly on the quantization step, with different optimal choices for each modality. The repeatability of PET and low-dose CT features should be carefully taken into account when selecting metrics to build multiparametric models.

**Key words:** PET/CT, texture analysis, radiomics, repeatability

## INTRODUCTION

The crucial role of positron emission tomography/computed tomography (PET/CT) with fluorine-18 fluorodeoxyglucose (FDG) for diagnosis and staging of non-small cell lung cancer (NSCLC) is established *(1)*. Tumor metabolism is usually quantified with standardized uptake value (SUV) metrics (*e.g.,* maximum and mean) in PET, whereas the low-dose CT component's role is limited to PET attenuation correction and anatomical localization.

*Radiomics* denotes the extraction of intensity, shape and heterogeneity features from medical images *(2)*. Its application to PET *(3)* and CT *(4)* has gained interest for characterizing NSCLC tumors quantitatively, with potentially higher value than standard metrics, with the opportunity to combine features from both PET and low-dose CT components *(5)*.

A first challenge is that numerous features can be calculated, most of which are sensitive to image noise, segmentation or reconstruction settings *(7–11)*. Their use for therapy response monitoring and early prediction faces another challenge: repeatability. Because metrics calculated in pre-, mid- and post-therapy images need to be compared, test-retest repeatability allows determining the cut-off above which a change is attributed to response or progression. This has been estimated at ±15% to 30% for SUV and volume *(12,13)*. Regarding shape and heterogeneity metrics, several studies have investigated their repeatability in PET with FDG or fluorine-18 fluorothymidine *(8,14–17)* and in diagnostic CT *(18,19)*, dosimetry CT *(4,18)*, contrast-enhanced CT (CE-CT) *(18,20)* or cone-beam CT (CBCT) *(21)*. These studies exploited small single-center cohorts [n=8 CE-CT *(20)*, n=10 CBCT *(21)*, n=11 FDG-PET *(8,15,17)*, n=11 fluorine-18 fluorothymidine-PET *(16)*, n=16 FDG-PET *(14)*, n=20 CT and 13 CE-CT *(18)* and n=31 CT *(4,19)*] and never reported on the repeatability of features from the low-dose CT from PET/CT, which is important when combining features from both components *(5)*.

Finally, it has been shown recently that the image quantization step in the calculation of textural features can have an impact on the relationship to other parameters *(3)* and on the repeatability *(17,22)*.

The primary goal of the present work was to evaluate the repeatability of shape and heterogeneity metrics from both PET and low-dose CT components in a large prospective multi-center cohort. A secondary goal was to evaluate the impact of the quantization step.

**MATERIALS AND METHODS**

**Patient cohort and imaging**

Patients with stage IIIB-IV NSCLC were prospectively included in the multi-center Merck MK-0646-008 (40 patients in 17 sites) and American College of Radiology Imaging Network (ACRIN) 6678 (34 patients in 14 sites) trials (NCT00424138 and NCT00729742, respectively) *(23)*. Centers had to conform to the criteria of ACRIN PET qualification (www.acrin.org/6678_protocol.aspx) to participate. Merck used a similar accreditation program. PET/CT protocols were designed in accordance with National Cancer Institute guidelines *(24)*. The institutional review board of each participating site approved the study, and all subjects signed a written informed consent form. The whole cohort of 74 patients has been previously included in *(23)*, but only SUV measurements were analyzed whereas in this present analysis, texture features and shape parameters were also computed both on PET and CT images. The present secondary analysis of deidentified PET/CT images from these trials was approved by ACRIN and was performed in compliance with the Health Insurance Portability and Accountability Act.

**PET and CT analysis**

In both test-retest datasets, the PET and the low-dose CT images were processed independently. In PET, the metabolically active volumes (MAV) of the primary tumor and up to three additional lesions were segmented with the Fuzzy Locally Adaptive Bayesian algorithm previously validated for accuracy and robustness *(25,26)*. In low-dose CT, the anatomical volume (AV) of primary tumors were delineated with a validated semi-automatic approach using 3D Slicer$^{TM}$ *(27)*. Additional lesions were analyzed if they could be reliably delineated.

The following metrics were calculated on the delineated volumes. Table 1 contains a glossary. All features are described with their calculation formulae *(3)* in the Supplemental Material.

3D shape descriptors were included, such as sphericity, irregularity or major axis *(4,28)*.

$1^{st}$-order metrics (not accounting for spatial distribution of voxels) in both Hounsfield units (low-dose CT) and SUV (PET) include maximum and mean values, as well as histogram-derived skewness, kurtosis, energy, entropy$_{HIST}$ or the area under the curve of the cumulative histogram (CH$_{AUC}$) *(29)*. These metrics do not require quantization as a prior step. Quantization (not to be confused with quantification) is an intensity resampling step applied to the image prior to building textures matrices on which $2^{nd}$ and $3^{rd}$ order features rely. These matrices dimensions are determined by the number of intensity values obtained after this resampling. Several different quantization approaches have been proposed *(3)*.

2nd-order metrics from grey-level co-occurrence matrix (GLCM) and neighborhood grey-tone difference matrix (NGTDM), and 3rd-order metrics from grey-level zone size matrix were calculated in a single matrix considering all 13 orientations simultaneously *(30,31)*. Quantization was performed in a set number of bins B (denoted from here onwards quantization$_B$), as previously recommended *(14,18,30,32)* using equation 1:

$$I_B = B \times \frac{I - I_{min}}{I_{max} - I_{min}} \quad (1)$$

Where $I_{max}$ and $I_{min}$ denote maximum and minimum intensity (Hounsfield units in low-dose CT and SUV in PET), and B is the number of bins (here B=64). Choosing a different B value can have an impact on the repeatability of features *(14)*. Results obtained with B=8 to 128 are in the Supplemental Material. It has been suggested that an alternative quantization using fixed-width bins (*e.g.,* 0.5 SUV) can have an important impact *(17,22)*. Results using this approach (denoted from here onwards quantization$_w$) following equation 2 were also generated.

$$I_W = \left\lceil \frac{I_O}{W} \right\rceil - \min\left(\left\lceil \frac{I_O}{W} \right\rceil\right) + 1 \quad (2)$$

Where W is the bin width (here 0.5 SUV for PET *(22)* and 10 Hounsfield units for low-dose CT). Note that W=0.25 SUV and W=5 Hounsfield units were also tested but no significant differences were observed. Supplemental Figure 1 shows a NSCLC tumor with both PET and low-dose CT, and the corresponding quantization results and histograms.

**Statistical analysis**

Statistical analyses were performed with MedCalc™ (MedCalc Software, Belgium). The repeatability of each metric was assessed with Bland-Altman analysis by reporting the mean and standard deviation (SD) of the differences between the two measurements. Lower and upper repeatability limits were calculated as ±1.96×SD after log-transformation when not normal. Bland-Altman analysis was preferred over intra-class correlation coefficients based on previous recommendations *(33)*. Intra-class correlation coefficients are nonetheless provided in the Supplemental Material. Correlations between metrics were assessed with Spearman rank coefficients (rs). Each metric was also categorized with respect to the repeatability (SD) of the corresponding volume of interest ($VOI_{rep}SD$): very reliable (≤0.5×$VOI_{rep}SD$), reliable (>0.5×$VOI_{rep}SD$ and ≤1.5×$VOI_{rep}SD$), moderately reliable (>1.5×$VOI_{rep}SD$ and ≤2×$VOI_{rep}SD$) and poorly reliable (>2×$VOI_{rep}SD$).

**RESULTS**

The analysis was performed in 73 datasets because one was not available. In the PET images, 73 primary tumors and 32 additional lesions (nodal or distant metastases) were analyzed. Mean MAV was 47.8 $cm^3$ (median 24.9 $cm^3$, SD 55.4 $cm^3$). In the low-dose CT, 2 patients were excluded because visual assessment of images indicated that repeatable volume delineation could not be ensured (Supplemental Fig. 2). Seventy-one primary tumors and 5 additional lesions were analyzed. Mean AV was 52.4 $cm^3$ (median 37.5 $cm^3$, SD 53.0 $cm^3$). .

Figure 1 displays repeatability results of volume determination in both modalities, while Figures 2, 3 and 4 display repeatability of 1st-order metrics and shape descriptors, 2nd- and 3rd-order textural features, respectively. Tables containing all results with also other quantization values are in the Supplemental Material.

**PET and low-dose CT volumes**

As shown in Figure 1, MAV determination had a repeatability of -1.4±11.1%, with upper and lower repeatability limits of +20.3% and -23.2%, which was dependent on MAV, smaller volumes exhibiting significantly (rs=-0.41, p<0.0001) poorer repeatability. The AV determination had a similar repeatability of -0.4±10.5%, with upper and lower repeatability limits of +20.3% and -21.0%. Repeatability was less dependent on volume (rs=-0.32, p=0.006).

PET (respectively low-dose CT) features were thus categorized with similar thresholds for reliability: ≤5.6% (respectively 5.3%), >5.6% (respectively 5.3%) and ≤16.7% (respectively 15.8%), >16.7% (respectively 15.8%) and ≤22.2% (respectively 21%) and >22.2% (respectively 21.0%).

**PET features**

**Shape descriptors and 1$^{st}$-order metrics**

Overall, the shape features in PET were very repeatable (Fig. 2). Irregularity and sphericity were very reliable, with only 4.8% SD. 3D surface and major axis were reliable although with higher variability (9.0% and 8.4%, respectively). Amongst intensity-based 1$^{st}$-order features, the most repeatable were CH$_{AUC}$ (-0.2 ± 3.6%) and entropy$_{HIST}$ (-0.2 ± 3.6%), whereas the least repeatable were energy (-1.2 ± 23.8%) and skewness (-1.1 ± 33.7%). Mean (SUV$_{mean}$) and max (SUV$_{max}$) values were moderately reliable, with upper and lower repeatability limits of -30.4% and 36.3%, and -34.3% and 41.3%, respectively.

**2$^{nd}$-order metrics**

As shown in Figure 3, with quantization$_B$, amongst GLCM features, entropy$_{GLCM}$ (-0.1 ± 2.6%), sum entropy (-0.2 ± 2.1%) and difference entropy (-0.2 ± 3.0%) were the most repeatable, whereas most other features fell in the reliable category. Five were categorized as moderately reliable and 3 as unreliable. For correlation the very poor repeatability is due to a few outliers for values around zero, to which Bland-Altman is very sensitive. After excluding them, correlation had reproducibility limits below ±20% and could be re-categorized as moderately reliable. The five NGTDM features were less repeatable than the best GLCM features although still categorized as reliable, all achieving SD ~14-17%, except contrast$_{NGTDM}$ (27.6%).

The use of the alternate quantization$_W$ changed both the above hierarchy and the absolute repeatability of the features. Overall, features calculated after quantization$_W$ were much less reliable with notably more outliers, all exhibiting a higher variability than MAV.

**3$^{nd}$-order metrics**

As shown in figure 4, amongst 3$^{rd}$-order metrics, quantization had a similar impact: with quantization$_W$ all grey-level zone size matrix features were categorized as poorly reliable, whereas with quantization$_B$ two were very reliable (small zone size emphasis and zone size percentage with SD <4%) and 3 reliable (large zone size emphasis, gray-level non-uniformity and zone size non-uniformity with SD ~11-14%). Amongst the least repeatable features were those focusing on small zones and/or low grey values (*e.g.*, LZLGE, SZLGE and LGLZE).

**Low-dose CT features**

**Shape descriptors and 1$^{st}$-order metrics**

As shown in Figure 2, morphological irregularity, sphericity and 3D surface were the most repeatable (SD 3.3%, 10.0% and 11.6%, respectively). Major axis was less reliable (3.8 ± 18.4%).

On the one hand, four histogram metrics showed poor reliability such as maximum (4.7 ± 38.6%) and mean (-4.2 ± 43.6%) intensity, kurtosis (4.8 ± 37.4%) and skewness (11.1 ± 202.2%). On the other hand, entropy$_{HIST}$ and CH$_{AUC}$ were very reliable (-0.1 ± 2.5% and 0.7 ± 9.1%).

**2$^{nd}$-order metrics**

The repeatability depended strongly on the quantization, quantization$_w$ improving the repeatability compared to quantization$_B$ (Fig. 3). Amongst GLCM metrics, the most repeatable (for quantization$_B$ *vs.* quantization$_w$, respectively) were entropy$_{GLCM}$ (-1.9 ± 12.0% *vs.* -0.4 ± 5.2%), sum entropy (-1.4 ± 10.0% *vs.* 0.1 ± 0.4%) and difference entropy (-2.3 ± 13.1% *vs.* -0.3 ± 1.9%). To a lesser extent, the same was observed for NGTDM, with higher repeatability using quantization$_w$. Complexity was the only parameter with variability <15.8% and categorized as reliable (0.5 ± 14.3% and -0.5 ± 12.3% with quantization$_B$ and quantization$_w$, respectively).

**3$^{nd}$-order metrics**

The quantization method also had an important impact (Fig. 4). Eight parameters were categorized as moderately reliable or better with quantization$_w$ and only two with quantization$_B$. Small zone size emphasis (-0.6 ± 4.8% *vs.* -0.5 ± 2.6% with quantization$_B$ and quantization$_w$, respectively) and zone size emphasis (-2.8 ± 17.4% *vs.* -0.9 ± 11.9%) were the most repeatable features (Figs. 4D and 4E).

**Impact of quantization method**

Overall, the inverted impacts of the quantization method observed in PET and low-dose CT can be explained by the different correlative relationships between the features and the corresponding volume and maximum intensity. In PET, we observed that quantization$_W$ features were correlated with SUV$_{max}$ and not with MAV. On the contrary, features calculated with quantization$_B$ were correlated with MAV but not SUV$_{max}$. The higher repeatability obtained with quantization$_B$ can thus be explained by the fact that MAV repeatability was much higher than that of SUV$_{max}$. Contrary to PET, features in low-dose CT were correlated with both volume and maximum intensity using quantization$_B$, whereas they were less or not correlated with either volume or intensity using quantization$_W$. Because maximum intensity had a much worse repeatability than volume in CT, quantization$_B$ thus led to worse repeatability. This is illustrated in Figure 5 for the feature dissimilarity. Note the relative inversion of relationships with volume and SUV$_{max}$ for quantization$_B$ compared to quantization$_W$ in the case of the PET component. On the contrary for the low-dose CT component, quantization$_B$ led to a higher correlation with maximum intensity than volume, but quantization$_W$ led to lower correlation with volume and non-significant correlation with maximum intensity.

**DISCUSSION**

In the present work, 73 test-retest PET/CT acquisitions from 31 centers (17 for ACRIN in the USA and 14 for Merck in Asia and Europe) were analyzed for repeatability.

A similar variability of volume delineations was observed for both modalities. MAV from PET were slightly smaller than AV measured in CT, mostly due to the fact that more lymph nodes and metastases were delineated in PET than in CT, and some large CT volumes had parts without FDG uptake. Regarding $SUV_{mean}$ and $SUV_{max}$, our results differ slightly from those previously published in the same cohort *(23)*. Only lesions with $SUV_{max}>4$ were included in the previous analysis, whereas we did not restrict it. By restricting to $SUV_{max}>4$, our test-retest results for $SUV_{max}$ were similar to those previously reported.

Regarding shape and heterogeneity features, our results confirm prior findings in PET *(8,14–17)*. To the best of our knowledge, our study is the first to report on the repeatability of these features in the low-dose CT component.

Overall, the geometric features (shape descriptors) were found reliable (some with high repeatability) in both modalities, which can be related to the high repeatability of segmentation. This is in line with previous findings for PET *(8,17)* and with morphological shape in other CT modalities *(4)*. We emphasize that only one segmentation by one expert was considered. The variability might be higher when considering different segmentation approaches and/or several observers.

Regarding $1^{st}$-order metrics and textural higher-order features, our results confirm that the repeatability varies greatly amongst metrics. On the one hand, several features were confirmed to be unreliable in both modalities and should be systematically avoided, *e.g.,* $1^{st}$-order skewness, $2^{nd}$-order Angular Second Moment, $contrast_{GLCM}$ and $contrast_{NGTDM}$, and $3^{rd}$-order metrics quantifying low grey values and/or small zones. On the other hand, it should be emphasized that several features were identified as reliable, in all three categories and for both modalities. In between, other features with moderate repeatability should be used with caution as they exhibit larger variability than the corresponding volume determination.

We compared two different quantization methods. Quantization$_B$ is most often used. The impact of choosing another B value has been evaluated previously *(14)* and our results confirm these findings. Although B=64 is a good compromise and most features exhibited similar repeatability with different values, repeatability of some metrics depended on B. We observed a different impact in PET and low-dose CT for quantization$_W$, as it led to worse repeatability in PET but better repeatability in low-dose CT. This was explained by the different relationships between the features and the corresponding volume and maximum intensity. With more control over data acquisition and higher repeatability of SUV$_{max}$, quantization$_W$ may lead to higher repeatability. These results highlight the major impact of the quantization step and its variable impact depending on image modality that should thus not be overlooked.

Our results confirm that studies building clinical models by combining features from PET/CT images should carefully account for repeatability. This is mandatory when calculating evolution of features across pre-, mid- and/or post-therapy images. This is nonetheless an important factor when building models based on single time-point images, as models built using robust and repeatable features are more likely to be generalizable and achieve good performance in external/testing cohorts. Repeatability is not the only criterion on which feature selection needs to be based, as discriminative power, robustness and redundancy have to be considered also.

Our study has limitations. Low-dose CT and PET images were analyzed separately using different segmentation processes performed independently on the test and re-test images. The repeatability evaluation therefore includes the intrinsic repeatability of the segmentation. We used robust segmentation approaches that should minimize variability. Another approach would consist in defining the volume on the test image and register it on the re-test image, which however requires accurate registration and raises other issues *(34)*. In a clinical environment, the use of less accurate and less robust segmentation could lead to a lower repeatability, especially for volume-correlated features.

We chose to categorize the repeatability levels of each metric with respect to that of the corresponding volume. The repeatability acceptance was similar for both modalities (reliability in PET was defined as SD below 16.5%, compared to 15.8% for low-dose CT). These thresholds are arbitrary and choosing different values would change the categorization of several metrics, but without changing their hierarchy.

Finally, respiratory gating was not applied. In NSCLC this may lead to different levels of quantitative bias between the test and retest images, as well as between PET and low-dose CT. The repeatability we reported are therefore larger than what could ideally be obtained in other body regions where motion is less important, or if respiratory motion correction was applied *(35)*.

**CONCLUSION**

Test-retest repeatability of shape and heterogeneity features in both components of PET/CT varied greatly amongst metrics. The repeatability also depended on the quantization step, with different optimal choices for PET or low-dose CT, because of different relationships of the metrics with volume or intensity. The repeatability of PET/CT features should be carefully accounted for when choosing metrics to combine in multiparametric models.

# REFERENCES

1. Sauter AW, Schwenzer N, Divine MR, Pichler BJ, Pfannenberg C. Image-derived biomarkers and multimodal imaging strategies for lung cancer management. *Eur J Nucl Med Mol Imaging*. 2015;4:634–643.

2. Lambin P, Rios-Velazquez E, Leijenaar R, et al. Radiomics: extracting more information from medical images using advanced feature analysis. *Eur J Cancer*. 2012;4:441–446.

3. Hatt M, Tixier F, Pierce L, Kinahan P, Cheze Le Rest C, Visvikis D. Characterization of PET images using texture analysis: the past, the present… any future? *Eur J Nucl Med Mol Imaging*. 2016:in press.

4. Aerts HJWL, Velazquez ER, Leijenaar RTH, et al. Decoding tumour phenotype by noninvasive imaging using a quantitative radiomics approach. *Nat Commun*. 2014;5:4006.

5. Desseroit M-C, D. Visvikis, Tixier F, et al. Development of a nomogram combining clinical staging with 18F-FDG PET/CT image features in Non-Small Cell Lung Cancer stage I-III. *Eur J Nucl Med Mol Imaging*. 2016;43:1477–1485.

6. Vaidya M, Creach KM, Frye J, Dehdashti F, Bradley JD, El Naqa I. Combined PET/CT image characteristics for radiotherapy tumor response in lung cancer. *Radiother Oncol J Eur Soc Ther Radiol Oncol*. 2012;102:239–245.

7. Hatt M, Tixier F, Cheze Le Rest C, Pradier O, Visvikis D. Robustness of intratumour $^{18}$F-FDG PET uptake heterogeneity quantification for therapy response prediction in oesophageal carcinoma. *Eur J Nucl Med Mol Imaging*. 2013;40:1662–1671.

8. Leijenaar RTH, Carvalho S, Velazquez ER, et al. Stability of FDG-PET Radiomics features: an integrated analysis of test-retest and inter-observer variability. *Acta Oncol Stockh Swed*. 2013;52:1391–1397.

9. Doumou G, Siddique M, Tsoumpas C, Goh V, Cook GJ. The precision of textural analysis in (18)F-FDG-PET scans of oesophageal cancer. *Eur Radiol*. 2015;25:2805–2812.

10. Galavis PE, Hollensen C, Jallow N, Paliwal B, Jeraj R. Variability of textural features in FDG PET images due to different acquisition modes and reconstruction parameters. *Acta Oncol*. 2010;49:1012–1016.

11. Yan J, Chu-Shern JL, Loi HY, et al. Impact of image reconstruction settings on texture features in 18F-FDG PET. *J Nucl Med*. 2015;56:1667–1673.

12. Wahl RL, Jacene H, Kasamon Y, Lodge MA. From RECIST to PERCIST: evolving considerations for PET response criteria in solid tumors. *J Nucl Med*. 2009;50 Suppl 1:122S–50S.

13. Hatt M, Cheze-Le Rest C, Aboagye EO, et al. Reproducibility of 18F-FDG and 3'-deoxy-3'-18F-fluorothymidine PET tumor volume measurements. *J Nucl Med.* 2010;51:1368–1376.

14. Tixier F, Hatt M, Le Rest CC, Le Pogam A, Corcos L, Visvikis D. Reproducibility of tumor uptake heterogeneity characterization through textural feature analysis in 18F-FDG PET. *J Nucl Med.* 2012;53:693–700.

15. Van Velden FHP, Nissen IA, Jongsma F, et al. Test-retest variability of various quantitative measures to characterize tracer uptake and/or tracer uptake heterogeneity in metastasized liver for patients with colorectal carcinoma. *Mol Imaging Biol MIB Off Publ Acad Mol Imaging.* 2014;16:13–18.

16. Willaime JMY, Turkheimer FE, Kenny LM, Aboagye EO. Quantification of intra-tumour cell proliferation heterogeneity using imaging descriptors of 18F fluorothymidine-positron emission tomography. *Phys Med Biol.* 2013;58:187–203.

17. Van Velden FHP, Kramer GM, Frings V, et al. Repeatability of Radiomic features in non-small-cell lung cancer [(18)F]FDG-PET/CT studies: impact of reconstruction and delineation. *Mol Imaging Biol MIB Off Publ Acad Mol Imaging.* 2016.

18. Fried DV, Tucker SL, Zhou S, et al. Prognostic value and reproducibility of pretreatment CT texture features in stage III non-small cell lung cancer. *Int J Radiat Oncol Biol Phys.* 2014;90:834–842.

19. Balagurunathan Y, Gu Y, Wang H, et al. Reproducibility and prognosis of quantitative features extracted from CT Images. *Transl Oncol.* 2014;7:72–87.

20. Yang J, Zhang L, Fave XJ, et al. Uncertainty analysis of quantitative imaging features extracted from contrast-enhanced CT in lung tumors. *Comput Med Imaging Graph Off J Comput Med Imaging Soc.* 2015;48:1–8.

21. Fave X, Mackin D, Yang J, et al. Can radiomics features be reproducibly measured from CBCT images for patients with non-small cell lung cancer? *Med Phys.* 2015;42:6784.

22. Leijenaar RTH, Nalbantov G, Carvalho S, et al. The effect of SUV discretization in quantitative FDG-PET Radiomics: the need for standardized methodology in tumor texture analysis. *Sci Rep.* 2015;5:11075.

23. Weber WA, Gatsonis CA, Mozley PD, et al., ACRIN 6678 Research team, MK-0646-008 Research team. Repeatability of 18F-FDG PET/CT in advanced non-small cell lung cancer: prospective assessment in 2 multicenter trials. *J Nucl Med Off Publ Soc Nucl Med.* 2015;56:1137–1143.

24. Shankar LK, Hoffman JM, Bacharach S, et al., National Cancer Institute. Consensus recommendations for the use of 18F-FDG PET as an indicator of therapeutic response in patients in National Cancer Institute trials. *J Nucl Med Off Publ Soc Nucl Med.* 2006;47:1059–1066.

25. Hatt M, Cheze le Rest C, Descourt P, et al. Accurate automatic delineation of heterogeneous functional volumes in positron emission tomography for oncology applications. *Int J Radiat Oncol Biol Phys*. 2010;77:301–308.

26. Hatt M, Cheze Le Rest C, Albarghach N, Pradier O, Visvikis D. PET functional volume delineation: a robustness and repeatability study. *Eur J Nucl Med Mol Imaging*. 2011;38:663–672.

27. Velazquez ER, Parmar C, Jermoumi M, et al. Volumetric CT-based segmentation of NSCLC using 3D-Slicer. *Sci Rep*. 2013;3.

28. Apostolova I, Rogasch J, Buchert R, et al. Quantitative assessment of the asphericity of pretherapeutic FDG uptake as an independent predictor of outcome in NSCLC. *BMC Cancer*. 2014;14:896.

29. Van Velden FH, Cheebsumon P, Yaqub M, et al. Evaluation of a cumulative SUV-volume histogram method for parameterizing heterogeneous intratumoural FDG uptake in non-small cell lung cancer PET studies. *Eur J Nucl Med Mol Imaging*. 2011;38:1636–1647.

30. Hatt M, Majdoub M, Vallières M, et al. 18F-FDG PET uptake characterization through texture analysis: investigating the complementary nature of heterogeneity and functional tumor volume in a multi-cancer site patient cohort. *J Nucl Med Off Publ Soc Nucl Med*. 2015;56:38–44.

31. Vallières M, Freeman CR, Skamene SR, El Naqa I. A radiomics model from joint FDG-PET and MRI texture features for the prediction of lung metastases in soft-tissue sarcomas of the extremities. *Phys Med Biol*. 2015;60:5471–5496.

32. Hunter LA, Krafft S, Stingo F, et al. High quality machine-robust image features: Identification in nonsmall cell lung cancer computed tomography images. *Med Phys*. 2013;40:121916.

33. Zaki R, Bulgiba A, Ismail R, Ismail NA. Statistical methods used to test for agreement of medical instruments measuring continuous variables in method comparison studies: a systematic review. *PloS One*. 2012;7:e37908.

34. Yip SSF, Coroller TP, Sanford NN, et al. Use of registration-based contour propagation in texture analysis for esophageal cancer pathologic response prediction. *Phys Med Biol*. 2016;61:906–922.

35. Yip S, McCall K, Aristophanous M, Chen AB, Aerts HJWL, Berbeco R. Comparison of texture features derived from static and respiratory-gated PET images in non-small cell lung cancer. *PloS One*. 2014;9:e115510.

**Table 1. Glossary**

| | |
|---|---|
| MAV | Metabolically active volume (PET) |
| AV | Anatomical volume (low-dose CT) |
| $CH_{AUC}$ | Area Under the Curve of the Cumulative Histogram |
| ASM | Angular Secondary Moment |
| IDM | Inverse Different Moment |
| ID | Inverse Difference |
| SOSV | Sum Of Square Variance |
| SAVE | Sum AVErage |
| SVAR | Sum VARiance |
| SENT | Sum ENTropy |
| DVAR | Difference VARiance |
| DENT | Difference ENTropy |
| IC | Information Correlation |
| TS | Texture strength |
| CP | Cluster Prominence |
| SZSE | Small Zone Size Emphasis |
| LZSE | Large Zone Size Emphasis |
| ZSNU | Zone Size Non-Uniformity |
| GLNU | Gray-Level Non-Uniformity |
| ZSP | Zone Size Percentage |
| LGLZE | Low Grey Level Zone Emphasis |
| HGLZE | High Grey Level Zone Emphasis |
| SZLGE | Small Zone / Low Grey Emphasis |
| SZHGE | Small Zone / High Grey Emphasis |
| LZLGE | Large Zone / Low Grey Emphasis |
| LZHGE | Large Zone High Grey Emphasis |

# FIGURE CAPTIONS

**Figure 1:** Bland-Altman analysis and correlation between volume and repeatability for MAV and AV determination.

**Figure 2**: Repeatability of $1^{st}$-order metrics and 3D shape descriptors measured on FDG PET (left) and low-dose CT (right). Features are ranked from highest (left) to lowest (right) repeatability. VR = very reliable ; R = reliable ; MR = moderately reliable ; PR = poorly reliable.

**Figure 3**: Repeatability of $2^{nd}$-order metrics measured on FDG PET (first row) and low-dose CT (second row), using either quantization$_B$ (first column) or quantization$_W$ (second column). Features are ranked from highest (left) to lowest (right) repeatability. VR = very reliable ; R = reliable ; MR = moderately reliable ; PR = poorly reliable.

**Figure 4:** Repeatability of $3^{rd}$-order metrics measured on FDG PET (first row) and low-dose CT (second row), using either quantization$_B$ (first column) or quantization$_W$ (second column). Features are ranked from highest (left) to lowest (right) repeatability. VR = very reliable ; R = reliable ; MR = moderately reliable ; PR = poorly reliable.

**Figure 5:** Illustration of correlative relationships between a textural feature (dissimilarity from GLCM) and volume (first row) or maximum intensity (second row), in both PET (first column) and low-dose CT (second column) components, depending on the quantization approach.